\documentclass[letterpaper]{article} 
\usepackage{aaai25}  
\usepackage{times}  
\usepackage{helvet}  
\usepackage{courier}  
\usepackage[hyphens]{url}  
\usepackage{graphicx} 
\usepackage{natbib}  
\usepackage{caption} 
\usepackage{algorithm}
\usepackage{algorithmic}

\usepackage{amsmath}
\usepackage{amssymb}
\usepackage[capitalize]{cleveref}
\usepackage{newfloat}
\usepackage{listings}
\DeclareCaptionStyle{ruled}{labelfont=normalfont,labelsep=colon,strut=off} 
\floatstyle{ruled}
\newfloat{listing}{tb}{lst}{}
\floatname{listing}{Listing}
\newcommand\Cgt{\mathbf{C}_{\mathrm{gt}}}
\newcommand\Csum{\mathbf{C}}
\newcommand\Cint{\hat{\mathbf{C}}}
\newcommand\nd{\mathbf{n}^d}
\newcommand\nt{\mathbf{n}^t}
\newcommand\np{\mathbf{n}^p}
\newcommand\Td{T}
\newcommand\Tt{\hat{T}}
\newcommand\sg{\texttt{sg}}

\title{Normal-NeRF: Ambiguity-Robust Normal Estimation for Highly Reflective Scenes}
\author{
    Ji Shi, Xianghua Ying\thanks{Xianghua Ying is the corresponding author.}, Ruohao Guo, Bowei Xing, Wenzhen Yue
}
\affiliations{
    State Key Laboratory of General Artificial Intelligence\\
    School of Intelligence Science and Technology, Peking University\\

    sjj118@pku.edu.cn, xhying@pku.edu.cn
}

\begin{document}

\maketitle

\begin{abstract}
    Neural Radiance Fields (NeRF) often struggle with reconstructing and rendering highly reflective scenes. Recent advancements have developed various reflection-aware appearance models to enhance NeRF's capability to render specular reflections. However, the robust reconstruction of highly reflective scenes is still hindered by the inherent shape ambiguity on specular surfaces. Existing methods typically rely on additional geometry priors to regularize the shape prediction, but this can lead to oversmoothed geometry in complex scenes. Observing the critical role of surface normals in parameterizing reflections, we introduce a transmittance-gradient-based normal estimation technique that remains robust even under ambiguous shape conditions. Furthermore, we propose a dual activated densities module that effectively bridges the gap between smooth surface normals and sharp object boundaries. Combined with a reflection-aware appearance model, our proposed method achieves robust reconstruction and high-fidelity rendering of scenes featuring both highly specular reflections and intricate geometric structures. Extensive experiments demonstrate that our method outperforms existing state-of-the-art methods on various datasets.
\end{abstract}

%
\begin{links}
    \link{Code}{https://github.com/sjj118/Normal-NeRF}
\end{links}

\begin{figure*}[!t]
    \centering
    \includegraphics[width=\textwidth]{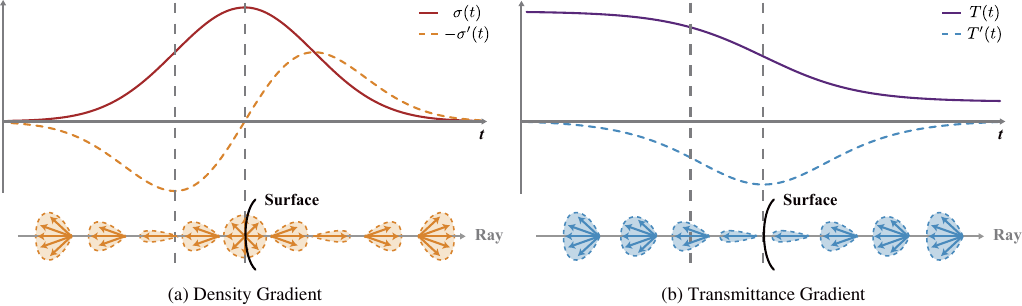}
    \caption{Illustration of our transmittance gradient compared to conventional density gradient for normal estimation. We plot the density $\sigma(t)$ and the transmittance $T(t)$ along a ray passing through a semi-transparent surface (top), and the distribution of the estimated normal vectors (bottom). (a) Since the derivatives of density approach zero near the surface, the directions of density gradients in nearby area can change rapidly. (b) In contrast, the derivatives of transmittance are large near the surface, thereby producing consistent normal estimates that align well with the true surface normals.}
    \label{fig:gradient}
\end{figure*}

\section{Introduction}
\label{sec:intro}

Neural Radiance Fields (NeRF) \cite{nerf} has been extensively studied for its powerful 3D scene reconstruction and rendering capabilities. NeRF utilizes multilayer perceptrons (MLPs) to encode a 3D scene into continuous fields of volume density and view-dependent radiance. Despite NeRF's proficiency in capturing fine geometric structures and smoothly varying view-dependent appearance, it often fails to accurately represent highly specular reflections, as the high-frequency view-dependent appearance prevents NeRF from eliminating the shape-radiance ambiguity \cite{nerf++}.

Recent studies attempt to enhance NeRF's capability in rendering reflective scenes by incorporating reflection-aware appearance models. Ref-NeRF \cite{ref-nerf} reparameterizes the view-dependent radiance as a function of the reflection direction about the surface normal, rather than the viewing direction itself. Later works further employ advanced reflection-aware appearance models, such as Microfacet Reflectance Model \cite{nmf} and Whitted-Style Ray Tracing \cite{mirror-nerf}. While these advanced reflection models are capable of representing complex reflections, achieving accurate reflection modeling remains a challenge due to the inherent shape ambiguity on highly reflective surfaces.

We observe that surface normals are critical in nearly all reflection-aware appearance models. Previous works typically derive surface normals based on the negative normalized gradient of the density field, which we term the ``density gradient''. However, this approach becomes unreliable on highly specular surfaces, as NeRF tends to fake specular reflections with foggy artifacts embedded behind the surfaces \cite{ref-nerf}. This phenomenon prevents the density values from monotonicly increasing towards the object's interior, resulting inconsistent gradient directions, as illustrated in Figure \ref{fig:gradient}(a).

In this paper, we propose an ambiguity-robust normal estimation pipeline to enhance NeRF's capability in reconstructing and rendering highly reflective scenes. To address the unreliability in conventional normal estimation approaches, we introduce the concept of ``transmittance gradient'', which can provide accurate normal estimates even under conditions of ambiguous shape predictions. Instead of calculating the gradient of the density, we compute the gradient of the transmittance, which is naturally monotonic along any specific ray, as illustrated in Figure \ref{fig:gradient}(b). Furthermore, we observe that reflection-aware appearance models generally prefer locally smooth surface normals, whereas NeRF tends to reconstruct a steep density field to achieve sharper object boundaries. To address this conflict, we propose a dual activated densities module that accommodates the distinct requirements of both the surface normals and the density field. Additionally, we design a stop-gradient warmup strategy for the predicted normal loss to prevent it from impeding the optimization of the density field.

By integrating a reflection-aware appearance model, our model achieves robust reconstruction and high-fidelity rendering in highly reflective scenes. Our key contributions can be summarized as follows:
\begin{itemize}
    \item To the best of our knowledge, we are the first to identify and analyze the inherent limitations of the  density gradient traditionally used for NeRF normal estimation.
    \item We introduce the \textbf{Transmittance Gradient} to address inconsistencies and irregularities in normal estimation within NeRF.
    \item We propose a \textbf{Dual Activated Densities} module to effectively bridge the gap between smooth surface normals and sharp object boundaries.
\end{itemize}

\section{Related Work}
\label{sec:related}

\subsubsection{Neural Radiance Fields.}

Neural Radiance Fields (NeRF) \cite{nerf} is a successful pipeline for novel view synthetic of complex scenes by optimizing the volumetric underlying function using a sparse set of input views. Later studies have improved the rendering quality \cite{mip-nerf,mip360,zipnerf} and accelerated the rendering speed \cite{instant-ngp,dvgo,tensorf} of NeRF using various techniques. NeRF has also inspired many subsequent works that extend its application, including few-shot rendering \cite{mvsnerf,pixelnerf,dietnerf,freenerf}, dynamic scene rendering \cite{nerfies,hypernerf,neural3dvideo}, and 3D generation \cite{graf,dreamfields,dreamfusion}.

\subsubsection{Radiance Fields with Reflections.}

Although the view-dependent radiance function of NeRF enables the modeling of non-Lambertian effects, it often encounters difficulties in accurately capturing the appearance of objects with high-frequency reflections due to the shape-radiance ambiguity \cite{nerf++}. NeRFReN \cite{nerfren} and MS-NeRF \cite{ms-nerf} treat specular reflections on planar mirrors as virtual images behind the surface, and model them with separate radiance fields to avoid inconsistency between front and back views of the mirror. Neural Catacaustics \cite{catacaustics} further leverages a neural warp field, enabling the modeling of reflections on non-planar surfaces. However, due to the lack of physically-based modeling of interactions between light and surfaces, these methods face challenges in accurately representing reflections on complex surfaces.

Recent works incorporate NeRF with reflection-aware appearance models. Ref-NeRF \cite{ref-nerf} conditions the view-dependent radiance on the reflected view direction instead of the camera view direction to make the radiance MLP easier to interpolate. Neural Microfacet Fields \cite{nmf} employs a microfacet reflectance model for physically-accurate reflection rendering. Mirror-NeRF \cite{mirror-nerf} models multi-bounce reflections with Whitted-Style Ray Tracing. ENVIDR \cite{envidr} leverages a pretrained neural renderer to enable high-fidelity scene relighting. SpecNeRF \cite{specnerf} proposes a Gaussian Directional Encoding to model near-field lighting in room-scale scenes. NeRF-Casting \cite{nerf-casting} efficiently casts reflection rays to synthesize consistent reflections.

\subsubsection{Normal Estimation within Radiance Fields.}

Estimating normals within NeRF poses a non-trivial challenge due to the absence of an explicit defined surface. Early research \cite{bi2020neural} employ an MLP to predict normal vectors directly without any regularization. Later works \cite{nerd,nerv} characterize the normal vectors as the negative normalized gradient of the density field, thereby enforcing the consistency between the surface normals and the density field. Subsequent studies  \cite{nerfactor,neroic,ref-nerf} integrate these approaches by tying the normals predicted by MLP to the normals estimated from the density field. However, these approaches become unreliable in highly reflective scenes, due to the non-monotonic nature of the density field under conditions of ambiguous shape predictions. More recent studies attempt to resolve this ambiguity by employing planar constraints \cite{mirror-nerf} or by enforcing the surfaces to be opaque \cite{nmf}. Nevertheless, these methods rely on additional geometry priors, which may not be suitable for all scenes.

Another series of research \cite{volsdf,neus} substitutes the density field in NeRF with a signed distance function (SDF), providing an explicit definition of surfaces. However, this approach also encounters challenges when reconstructing surfaces within highly reflective scenes. Ref-NeuS \cite{ref-neus} attempt to reduce the ambiguity by calculating a reflection score to identify specular regions. Despite this, errors in the reflection score can still lead to incorrect geometry reconstruction, particularly on concave surfaces.

\section{Preliminary}
\label{sec:pre}

Neural Radiance Fields (NeRF) \cite{nerf} encodes a scene as continuous volumetric fields, where the density $\sigma(\mathbf{x})\in \mathbb{R}$ at any 3D position $\mathbf{x}\in\mathbb{R}^3$ and the color $\mathbf{c}(\mathbf{x},\mathbf{d})\in\mathbb{R}^3$ at any 3D position $\mathbf{x}\in\mathbb{R}^3$ under any viewing direction $\mathbf{d}\in\mathbb{R}^2$ can be queried from MLPs. The color of a ray $\mathbf{r}(t)=\mathbf{o}+t\mathbf{d}$ is rendered as:
\begin{equation}
    \Cint(\mathbf{r})=\int_{0}^{+\infty}T(t)\sigma(\mathbf{r}(t))\mathbf{c}(\mathbf{r}(t),\mathbf{d})\mathrm{d}t\,,
    \label{eq:nerf_color_int}
\end{equation}
where
\begin{equation}
    \Td(t)=\exp{\left(-\int_{0}^{t}\sigma(\mathbf{r}(s))\mathrm{d}s\right)}
    \label{eq:nerf_trans_int}
\end{equation}
is the transmittance along the ray, which indicates the probability of light traveling along the ray over the interval $[0, t)$ without being absorbed or scattered. To approximate the integral $\Cint(\mathbf{r})$, NeRF samples a set of points $\{\mathbf{x}_i=\mathbf{o}+t_i\mathbf{d}\}$ and denotes the distance between adjacent samples by $\delta_i=t_{i+1}-t_{i}$. The density $\sigma_i$ and the color $\mathbf{c}_i$ at each point $\mathbf{x}_i$ under the direction $\mathbf{d}$ are then queried to approximate the color of the ray as:
\begin{equation}
    \Csum(\mathbf{r})=\sum_{i}\Td_i(1-\exp{(-\sigma_i\delta_i)})\mathbf{c}_i\,,
    \label{eq:nerf_color_sum}
\end{equation}
where
\begin{equation}
    \Td_i=\exp{\left(-\sum_{j<i}\sigma_i\delta_i\right)}\,.
    \label{eq:nerf_trans_sum}
\end{equation}

NeRF is optimized by minimizing the L2 difference between the ground truth color $\Cgt(\mathbf{r})$ of each pixel taken from input images and the predicted color $\Csum(\mathbf{r})$ of the ray corresponding to this pixel:
\begin{equation}
    \mathcal{L}_{\mathrm{c}}(\mathbf{r})=\left\|\Csum(\mathbf{r})-\Cgt(\mathbf{r})\right\|^2\,.
    \label{eq:color_loss}
\end{equation}

\section{Irregularity in Normal Estimation}
\label{sec:density_gradient}

Since volume density and surface normals both characterize object shape, it naturally follows to estimate normal vectors from the density field. Observing that volume density increases drastically at the boundary between non-opaque air and opaque objects, NeRD \cite{nerd} characterizes normal vectors directly as the negative normalized gradients of the density field, which we refer to as the ``density gradients'':

\begin{equation}
    \nd(\mathbf{x})=-\frac{\nabla\sigma(\mathbf{x})}{\|\nabla\sigma(\mathbf{x})\|}\,.
    \label{eq:density_grad}
\end{equation}

However, this approach is only effective on opaque surfaces, where the volume density can strictly increase towards the interior of the object. When encountering highly reflective surfaces, NeRF tends to fake reflections by positioning them underneath the surfaces, resulting in semi-transparent surface predictions during the optimization process. Local maxima in the density field will inevitably occur near a semi-transparent surface. Since the gradient at the maximum point is zero, the directions and magnitudes of gradients in nearby areas can change rapidly, leading to an irregular distribution of estimated normal vectors, as illustrated in Figure \ref{fig:gradient}(a). Additionally, the gradients on either side of the surface may point in opposite directions.

Later studies \cite{ref-nerf,nmf} introduce an orientation loss to penalize normal vectors that face away from the camera. However, this regularization encourages opaque surface predictions, compelling the model to select a specific surface from potential candidates and discard the rest without substantial evidence. Therefore, these methods still struggle with the accurate reconstruction of highly reflective scenes, as demonstrated in Figure \ref{fig:density-based}.

\section{Method}
\label{sec:method}

Our goal is to enhance NeRF's robustness and fidelity in reconstructing and rendering highly reflective scenes. We begins by introducing the concept of the transmittance gradient to address the irregularity in normal estimation under ambiguous shape conditions. Subsequently, we employ the dual activated densities to meet the distinct requirements of surface normals and object boundaries. Finally, we present details of our training process, including a stop-gradient warmup strategy for the predicted normal loss, and a reflection-aware appearance model.

\subsection{Transmittance Gradient}
\label{sec:method:approximation}

Despite the failure of density gradients to provide robust normal estimates on highly reflective surfaces, the density field remains a valuable source of geometric information for estimating normal vectors. Additionally, we find that the source of the irregularity in density gradients can be attributed to the non-monotonic nature of the volume density. Building on these insights, we identify the transmittance, as defined in Eq. \eqref{eq:nerf_trans_int}, as a promising mediator for linking the density field and surface normals. The accumulated transmittance along any specific ray is a monotonically decreasing function, and its derivative at any position is equal to the rendering weight at that position \cite{nerf_digest}. Therefore, any sample point that contributes significantly to the final rendering will possesses a correspondingly large derivative value, thereby avoiding the irregularity encountered by density gradients, as illustrated in Figure \ref{fig:gradient}(b).

However, Both the transmittance and its derivative are defined along a specific 1D ray. To extract 3D directional information from them, we introduce the concept of the ``transmittance gradient'' as the normalized gradient of the transmittance over a segment of ray with respect to the translation of this ray. More specifically, consider a ray originating at point $\mathbf{o}$ and directed along $\mathbf{d}$. For any point $\mathbf{x}=\mathbf{o}+t\mathbf{d}$ on this ray, the transmittance over the segment between $\mathbf{o}$ and $\mathbf{x}$ can be reparameterized by substituting $\mathbf{o}$ with $\mathbf{x}-t\mathbf{d}$:
\begin{equation}
    \Tt(\mathbf{x};\mathbf{d},t)=\exp{\left(-\int_{0}^{t}\sigma(\mathbf{x}-s\mathbf{d})\mathrm{d}s\right)}\,.
    \label{eq:out_trans_int}
\end{equation}
Then the transmittance gradient at point $\mathbf{x}$ is formulated as:
\begin{equation}
    \begin{aligned}
        \nt(\mathbf{x};\mathbf{d},t)=
        \frac{{\nabla_{\mathbf{x}}\Tt(\mathbf{x};\mathbf{d},t)}}{\left\|\nabla_{\mathbf{x}}\Tt(\mathbf{x};\mathbf{d},t)\right\|}\,.
        \label{eq:normal_int}
    \end{aligned}
\end{equation}
If the ray is originated from a camera and the point $\mathbf{x}$ is visible to the camera, the transmittance gradient $\nt(\mathbf{x};\mathbf{d},t)$ serves as an estimate of the normal vector at the position $\mathbf{x}$.

Unlike the density, which is encoded by a differentiable MLP, the transmittance in NeRF is approximated through numerical integration. Consequently, the transmittance gradient cannot be directly obtained via automatic differentiation. Following the quadrature rule used in NeRF, we estimate the transmittance gradient with the same discrete set of samples:
\begin{equation}
    \nt_i=-\frac{\sum_{j<i}\nabla\sigma(\mathbf{x}_j)\delta_j}{\left\|\sum_{j<i}\nabla\sigma(\mathbf{x}_j)\delta_j\right\|}\,.
    \label{eq:normal_sum}
\end{equation}

\subsection{Dual Activated Densities}
\label{sec:coactivate}

While our transmittance gradient effectively mitigates irregularities in normal estimation under conditions of ambiguous shape prediction, the disparity between the density and surface normal can still induce instability. Sharp object boundaries, which are crucial for high-fidelity renderings and geometric details \cite{dvgo}, necessitate a steep density field. In contrast, reflection-aware appearance models generally prefer locally smooth surface normals to accurately reconstruct high-frequency reflections. To bridge this gap, we propose a dual activated densities module that simultaneously supports smooth surface normals and maintains sharp object boundaries.

In our design, we apply two distinct activation functions, \texttt{softplus} and \texttt{exp}, to the output of the density MLP. Specifically, we activate a sharp density, $\hat{\sigma} = \exp(b)$, and a smooth density, $\tilde{\sigma} = \text{softplus}(b)$, where $b$ represents the pre-activation output of the density MLP. The sharp density $\hat\sigma$ will be used to calculate the rendering weights in \cref{eq:nerf_color_sum,eq:nerf_trans_sum}, while the smooth density $\tilde\sigma$ will be used to compute the transmittance gradient according to \cref{eq:normal_sum}, by substituting $\sigma$ in these formulas. Shared learnable MLP parameters ensure consistency between the two densities, while dual activation functions allow for varying degrees of steepness.

\begin{figure}[!b]
    \centering
    \includegraphics[width=\columnwidth]{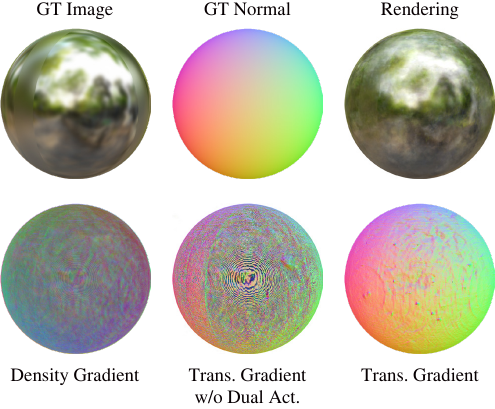}
    \caption{We intentionally reconstruct a highly reflective scene using a baseline model that excludes any reflection-aware appearance model, which is unable to eliminate the ambiguity in shape prediction. Under this ambiguous shape condition, we visually compare different normal estimation methods. The density gradient method completely fails to produce reasonable normal estimates. Omitting dual activated densities in our transmittance gradient method leads to artifacts from numerical instability, while our full pipeline produces normal estimates that align well with the ground truth.}
    \label{fig:normal}
\end{figure}

Moreover, the dual activated densities also reduces numerical instability when calculating the transmittance gradient. As our transmittance gradient is approximated through numerical integration, a steep density field can induce artifacts caused by numerical instability, as demonstrated in Figure \ref{fig:normal}. Conversely, the smooth density $\tilde\sigma$ activated by \texttt{softplus} significantly enhances the stability of the numerical integration process.

\begin{figure*}[t]
    \centering
    \includegraphics[width=\textwidth]{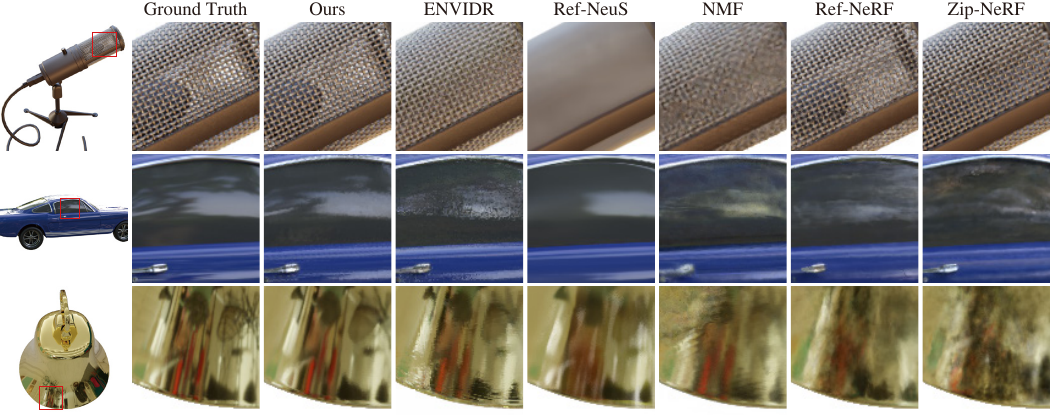}
    \caption{Qualitative comparisons on test views of synthetic scenes.}
    \label{fig:sota}
\end{figure*}

\subsection{Training}
\label{sec:method:normal_field}

While the transmittance gradient provides an accurate normal estimate, it lacks precision. Specifically, the transmittance gradient value at a given spatial position can vary slightly with changes in viewing direction and sampling strategy. To address this, we employ a spatial MLP to refine the transmittance gradient, ensuring both accuracy and precision in normal prediction, which is crucial for reflection modeling. For any position $\mathbf{x}$ within the scene, we predict the normal $\np(\mathbf{x})\in\mathbb{R}^3$ by normalizing a three-dimensional vector output from the spatial MLP.

\subsubsection{Predicted Normal Loss with Stop-Gradient Warmup.}

Ref-NeRF \cite{ref-nerf} ties the normals predicted by MLP and the normals estimated from the density field using a simple predicted normal loss. However, we observe that this loss function may impede the optimization of the density field. To address this issue, we extend the predicted normal loss with a stop-gradient warmup strategy. Specifically, we apply the stop-gradient operator $\sg$ to allow for the adjustment of the proportion of gradients flowing from the predicted normals $\np$ to the density field (including the rendering weights $w$ and the transmittance gradients $\nt$):
\begin{equation}
    \begin{aligned}
        \mathcal{L}_\mathrm{n} & =\lambda_\mathrm{n}\overleftrightarrow{\mathcal{L}_\mathrm{n}}+(1-\lambda_\mathrm{n})\overrightarrow{\mathcal{L}_\mathrm{n}}\,, \\
    \end{aligned}
\end{equation}
where
\begin{equation}
    \begin{aligned}
        \overleftrightarrow{\mathcal{L}_\mathrm{n}} & =\sum_{i}w_i\|\np_i-\nt_i\|^2 \,,                      \\
        \overrightarrow{\mathcal{L}_\mathrm{n}}     & =\sum_{i}\sg(w_i)\left\|\np_i-\sg(\nt_i)\right\|^2 \,. \\
    \end{aligned}
\end{equation}
In all of our experiments, the parameter $\lambda_\mathrm{n}$ follows an exponential warmup, increasing from $0.01$ to $1$ over 20k iterations.

Our key insight regarding this design is that the density field is more reliable than the predicted normals at the beginning of training. Although both the density field and predicted normals are initialized randomly, the density field quickly converges to reasonable shape predictions. Conversely, in the absence of the predicted normal loss, the predicted normals may even degenerate into a piecewise constant function \cite{ref-nerf}. Consequently, at the beginning of training, the gradients from the predicted normals could disrupt the density field. In contrast, the gradients from the density field can help in regularizing the predicted normals. Therefore, we restrict the majority of gradients flowing from the predicted normals to the density field at the beginning of training.

\begin{table*}[t]
    \setlength{\tabcolsep}{2mm}
    \centering
    \begin{tabular}{l|cccc|cccc|ccc}
        \hline
                 & \multicolumn{4}{c|}{NeRF Synthetic} & \multicolumn{4}{c|}{Shiny Blender} & \multicolumn{3}{c}{Glossy Synthetic}                                                                                                                                                                  \\
                 & PSNR$\uparrow$                      & SSIM$\uparrow$                     & LPIPS$\downarrow$                    & MAE$\downarrow$    & PSNR$\uparrow$    & SSIM$\uparrow$    & LPIPS$\downarrow$ & MAE$\downarrow$   & PSNR$\uparrow$    & SSIM$\uparrow$    & LPIPS$\downarrow$ \\
        \hline
        Zip-NeRF & $\mathit{33.69}$                    & \textbf{0.974}                     & \textbf{0.028}                       & -                  & 30.12             & 0.948             & 0.089             & -                 & 29.88             & 0.944             & $\mathit{0.065}$  \\
        Ref-NeRF & \underline{33.99}                   & $\mathit{0.966}$                   & $\mathit{0.038}$                     & 24.077             & \underline{35.96} & 0.967             & $\mathit{0.059}$  & 18.384            & $\mathit{30.39}$  & 0.941             & 0.074             \\
        NMF      & 30.71                               & 0.940                              & 0.053                                & $\mathit{20.952}$  & 34.56             & 0.963             & \underline{0.053} & 6.061             & 28.76             & 0.932             & 0.086             \\
        Ref-NeuS & 25.83                               & 0.916                              & 0.099                                & 22.973             & 33.26             & $\mathit{0.971}$  & \underline{0.053} & $\mathit{4.852}$  & \underline{30.49} & \underline{0.956} & \underline{0.064} \\
        ENVIDR   & 29.33                               & 0.942                              & 0.064                                & \underline{20.549} & $\mathit{35.02}$  & \underline{0.972} & \underline{0.053} & \underline{4.602} & 28.47             & $\mathit{0.948}$  & 0.066             \\
        Ours     & \textbf{34.53}                      & \underline{0.972}                  & \underline{0.030}                    & \textbf{14.486}    & \textbf{39.24}    & \textbf{0.982}    & \textbf{0.040}    & \textbf{4.241}    & \textbf{33.24}    & \textbf{0.971}    & \textbf{0.043}    \\
        \hline
    \end{tabular}
    \caption{Comparison with SOTAs on NeRF Synthetic \cite{nerf}, Shiny Blender \cite{ref-nerf} and Glossy Synthetic \cite{nero}. The best results are \textbf{bold}, the second best results are \underline{underlined}, and the third best results are \textit{italics}.}
    \label{tab:sota}
\end{table*}

\begin{figure}[t]
    \centering
    \includegraphics[width=\columnwidth]{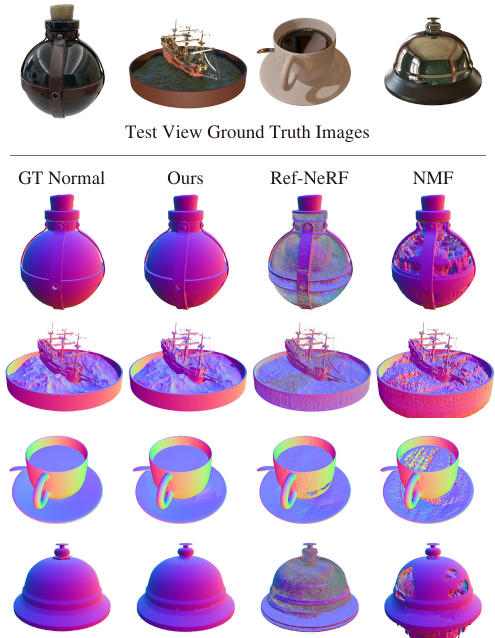}
    \caption{Normal map visualizations of NeRF-based methods.}
    \label{fig:density-based}
\end{figure}

\begin{figure}[t]
    \centering
    \includegraphics[width=\columnwidth]{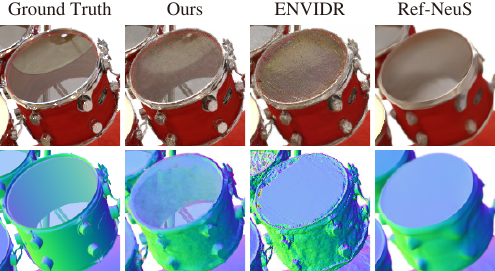}
    \caption{Visualization on a reflective yet semi-transparent surface.}
    \label{fig:sdf-based}
\end{figure}

\begin{table}[t]
    \centering

    \begin{tabular}{l|ccc}
        \hline
                 & PSNR$\uparrow$    & SSIM$\uparrow$    & LPIPS$\downarrow$ \\
        \hline
        Zip-NeRF & 23.44             & \underline{0.626} & \textbf{0.257}    \\
        Ref-NeRF & \underline{24.06} & 0.589             & 0.355             \\
        Ours     & \textbf{24.09}    & \textbf{0.630}    & \underline{0.293} \\
        \hline
    \end{tabular}
    \caption{Comparison with SOTAs on real captured scenes from Ref-NeRF \cite{ref-nerf}.}
    \label{tab:real}
\end{table}

\subsubsection{Reflection-Aware Appearance.}

To demonstrate the efficacy of our normal estimation technique, we develop our reflection-aware appearance model as follows. For any sample point $\mathbf{x}$ observed under viewing direction $\mathbf{d}$, we compute the reflection direction using the predicted normal $\np$ at this position:
\begin{equation}
    \mathbf{d}^r=2(\mathbf{d}\cdot\np)\np-\mathbf{d}\,,
\end{equation}

We then feed the reflection direction into an environment MLP $\mathcal{F}_\mathrm{env}$ to obtain an environment feature:
\begin{equation}
    \mathbf{f}_\mathrm{env}=\mathcal{F}_\mathrm{env}(\mathbf{d}^r)\,.
\end{equation}

Incorporating a material feature $\mathbf{f}_\mathrm{mat}$ conditioned exclusively on spatial position, we decompose the color into its diffuse and specular components:
\begin{equation}
    \begin{aligned}
         & \mathbf{c}_\mathrm{s}=\mathcal{F}_\mathrm{s}(\mathbf{f}_\mathrm{mat},\mathbf{f}_\mathrm{env})\,, \\
         & \mathbf{c}_\mathrm{d}=\mathcal{F}_\mathrm{d}(\mathbf{f}_\mathrm{mat})\,.
    \end{aligned}
\end{equation}

Finally, we combine the diffuse component and the specular component in linear space and then convert it to sRGB space with gamma tone mapping \cite{srgb}:
\begin{equation}
    \mathbf{c}=\gamma(\mathbf{c}_\mathrm{d}+\mathbf{c}_\mathrm{s})\,.
\end{equation}

\section{Experiments}
\label{sec:exp}

\subsubsection{Datasets.}
To comprehensively validate the effectiveness and robustness of our proposed method, we conduct evaluation on several datasets, including widely-used NeRF Synthetic dataset \cite{nerf}, two reflective objects datasets: Shiny Blender \cite{ref-nerf} and Glossy Synthetic \cite{nero}, and one real captured dataset from Ref-NeRF \cite{ref-nerf}.

\subsubsection{Baselines and Metrics.}
We compare our method against the following baselines: \textbf{Zip-NeRF} \cite{zipnerf}, a state-of-the-art grid-based NeRF variant with no special treatment for reflection; \textbf{Ref-NeRF} \cite{ref-nerf}, a NeRF-based method focusing on reflective objects rendering; \textbf{NMF} \cite{nmf}, a NeRF-based method for inverse rendering with microfacet reflectance model; \textbf{Ref-NeuS} \cite{ref-neus}, a SDF-based method for reflective surface reconstruction with a reflection-aware photometric loss. \textbf{ENVIDR} \cite{envidr}, a SDF-based method for scene relighting with a pretrained neural renderer.
We evaluate rendering quality using PSNR, SSIM and LPIPS \cite{lpips}, and assess normal accuracy with mean angular error (MAE) \cite{ref-nerf}.

\subsubsection{Implementation Details.}
All experiments are conducted on an NVIDIA RTX 4090 GPU. We implement our model within Nerfstudio \cite{nerfstudio}, building upon the Instant-NGP \cite{instant-ngp} framework. We train our model for 50k iterations with a batch size of $2^{19}$ sample points. Please refer to supplementary material for more details.

\begin{figure}[t]
    \centering
    \includegraphics[width=\columnwidth]{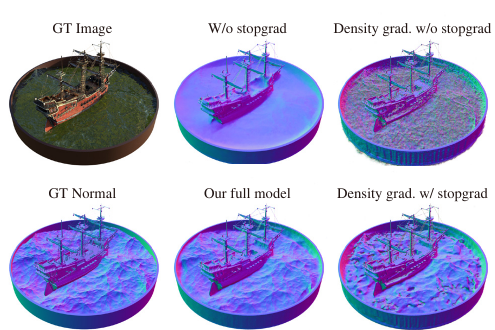}
    \caption{Ablation on normal estimation techniques, including the choice between transmittance gradient and density gradient, and the application of the stop-gradient warmup strategy.}
    \label{fig:ablation-ship}
\end{figure}

\subsection{Comparison with State-of-the-Arts}
\label{sec:exp:compare}
Table \ref{tab:sota} reports quantitative results on synthetic datasets. Our method demonstrates superior performance on Shiny Blender dataset and  Glossy Synthetic dataset, and achieves performance on par with Zip-NeRF \cite{zipnerf} on non-specular NeRF Synthetic dataset. Visual comparisons of rendering quality are demonstrated in Figure \ref{fig:sota}. Other NeRF-based baselines (Ref-NeRF, NMF and Zip-NeRF) struggle with the highly specular reflections (second and third rows), while SDF-based baselines (ENVIDR and Ref-NeuS) fails to capture the intricate geometric details (first row). Our method consistently recovers high-fidelity rendering across all these scenes, indicating the robustness and effectiveness of our method. Per-scene metrics and additional visualizations are presented in supplementary material.

We further visualize and compare the recovered normal maps of NeRF-based methods in Figure \ref{fig:density-based}. We can see that Ref-NeRF \cite{ref-nerf} tends to generate semi-transparent surfaces with noisy normal maps, due to its inability to sufficiently resolve ambiguities in highly reflective surfaces. In contrast, NMF \cite{nmf}, which employs the orientation loss directly to density gradients, recovers surfaces that are opaque but often irregular. Our model demonstrates a robust capability to produce accurate surface normals, effectively mitigating such ambiguities.

Figure \ref{fig:sdf-based} highlights a reflective yet semi-transparent surface, which SDF-based baselines reconstruct as opaque. In comparison, our model effectively preserves the semi-transparency of reconstructed surface while still generating a plausible normal map.

To explore our method's robustness in real world environments, we conduct experiments using the real captured scenes from Ref-NeRF \cite{ref-nerf}. The quantitative results presented in Table \ref{tab:real} show that our method performs on par with existing methods.

\begin{table}[t]
    \centering
    \setlength{\tabcolsep}{1mm}
    \begin{tabular}{cc|cc|ccc}
        \hline
        \multicolumn{2}{c}{Normal} & \multicolumn{2}{c}{Density} & \multicolumn{3}{c}{Metrics}                                                                          \\
        \hline
        Trans.                     & Stopgrad                    & Softplus                    & Exp        & PSNR$\uparrow$    & SSIM$\uparrow$    & LPIPS$\downarrow$ \\
        \hline
        \checkmark                 & \checkmark                  & \checkmark                  & \checkmark & \textbf{33.24}    & \textbf{0.971}    & \textbf{0.043}    \\
        \checkmark                 &                             & \checkmark                  & \checkmark & 31.64             & 0.950             & 0.072             \\
                                   & \checkmark                  & \checkmark                  & \checkmark & 30.67             & 0.957             & 0.062             \\
                                   &                             & \checkmark                  & \checkmark & 27.82             & 0.925             & 0.108             \\
        \hline
        \checkmark                 & \checkmark                  & \checkmark                  &            & $\mathit{32.20}$  & $\mathit{0.965}$  & $\mathit{0.052}$  \\
        \checkmark                 & \checkmark                  &                             & \checkmark & \underline{32.79} & \underline{0.968} & \underline{0.048} \\
        \hline
    \end{tabular}
    \caption{Quantitative comparisons for ablation runs on Glossy Synthetic dataset \cite{nero}.}
    \label{tab:ablation}
\end{table}

\begin{figure}[t]
    \centering
    \includegraphics[width=\columnwidth]{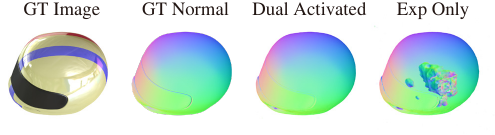}
    \caption{Normal map comparison between dual activated densities and single exp activated density at 10K iterations.}
    \label{fig:ablation-exp}
\end{figure}

\begin{figure}[!t]
    \centering
    \includegraphics[width=\columnwidth]{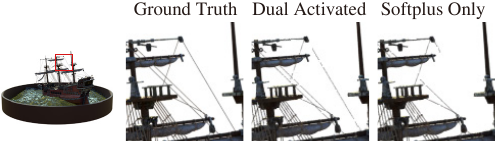}
    \caption{Visual comparison between dual activated densities and single softplus activated density.}
    \label{fig:ablation-softplus}
\end{figure}

\begin{table}[!t]
    \centering
    \begin{tabular}{l|cccc}
        \hline
        $\lambda_\mathrm{n}$ Scheduling & PSNR$\uparrow$    & SSIM$\uparrow$    & LPIPS$\downarrow$ & MAE$\downarrow$   \\
        \hline
        Exp Warmup                      & \textbf{37.02}    & \textbf{0.979}    & \textbf{0.048}    & \textbf{4.534}    \\
        Constant 0                      & 36.40             & $\mathit{0.976}$  & 0.054             & 6.143             \\
        Constant 0.01                   & \underline{36.82} & \underline{0.978} & $\mathit{0.051}$  & \underline{5.199} \\
        Constant 0.1                    & $\mathit{36.78}$  & \underline{0.978} & \underline{0.049} & $\mathit{5.353}$  \\
        Constant 1                      & 36.35             & $\mathit{0.976}$  & 0.052             & 5.767             \\
        \hline
    \end{tabular}
    \caption{Ablation on the scheduling of $\lambda_\mathrm{n}$, which controls the ratio of stop-gradient, conducted on Shiny Blender dataset \cite{ref-nerf}.}
    \label{tab:ablation-stopgrad}
\end{table}

\subsection{Ablation Studies}
\label{sec:exp:ablation}

We conduct a series of ablation studies to evaluate the effect of our key components.

\subsubsection{Normal Estimation.}
We evaluate both transmittance gradient and the conventional density gradient for normal estimation, each with and without the stop-gradient warmup strategy. Quantitative results in Table \ref{tab:ablation} demonstrate that the absence of either the transmittance gradient or the stop-gradient warmup strategy leads to a significant degradation in performance. Without the implementation of the stop-gradient warmup strategy, the randomly initialized predicted normals may oversmooth the surface reconstruction of water waves, as illustrated in Figure \ref{fig:ablation-ship}. Normal estimation based on the density gradients faces challenges in accurately reconstructing the water waves, even when employing the stop-gradient warmup strategy.

\subsubsection{Density Activation.}
In addition to the dual activated densities, we also evaluate the performance using only an exp or softplus density activation. As presented in Table \ref{tab:ablation}, employing a single activated density uniformly diminishes the metrics. As demonstrated in Figure \ref{fig:ablation-exp}, using a single exp activated density undermines the robustness of normal estimation. Additionally, employing a single softplus activated density hampers the reconstruction of thin geometric structures, as illustrated in Figure \ref{fig:ablation-softplus}.

\subsubsection{Stop-Gradient Warmup.}
The parameter $\lambda_\mathrm{n}$, which controls the stop-gradient ratio in the predicted normal loss, follows an exponential warmup schedule in our design. We compare its performance with that of a constant $\lambda_\mathrm{n}$ applied throughout training. Table \ref{tab:ablation-stopgrad} shows that our warmup strategy consistently outperforms all tested constant values of $\lambda_\mathrm{n}$.

\section{Conclusion}
\label{sec:conclusion}

In this paper, we present a pipeline to enhance NeRF's capability in reconstructing and rendering highly reflective scenes. The core of our approach is a transmittance-gradient-based normal estimation technique to improve the robustness and accuracy of surface normal estimation under conditions of ambiguous shape prediction. We also introduce dual activated densities to model objects with both smooth surfaces and sharp boundaries. Extensive experiments demonstrate that our approach quantitatively and qualitatively outperforms existing methods.

\section{Acknowledgments}
This work was supported by the National Natural Science Foundation of China (NSFC) under Grant No. 62371009, and Beijing Natural Science Foundation under Grant No. L247029.

\bibliography{aaai25}

\clearpage

\section{Implementation Details}
\begin{figure}[t]
    \centering
    \includegraphics[width=\columnwidth]{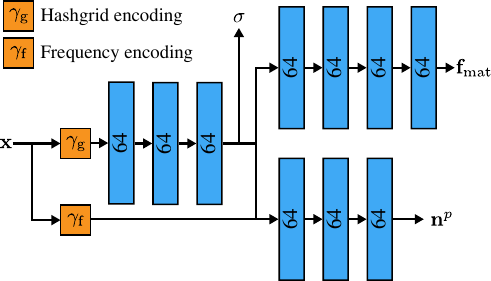}
    \caption{Architecture of spatial MLP. Dimension of each linear layer (illustrated as blue block) is $64$.}
    \label{fig:arch_geo}
\end{figure}

Our method is implemented within Nerfstudio framework \cite{nerfstudio}, based on Instant-NGP \cite{instant-ngp}. Each hidden layer in our MLPs is followed by a ReLU activation. The architecture of ``spatial MLP'' that predicts density $\sigma$, normal $\mathbf{n}^p$ and material feature $\mathbf{f}_{\mathrm{mat}}$ given any spatial location $\mathbf{x}$, is illustrated in Figure \ref{fig:arch_geo}. In the optimization of synthetic scenes, the hash grid positional encoding \cite{instant-ngp} $\gamma_{\mathrm{g}}$ has $16$ layers with resolutions ranging from $16$ to $2048$, with a hash table size of $2^{19}$ and feature dimension of $2$. For large scenes captured from real world, we expand the hash grid's resolutions to range from $16$ to $8192$, while increasing the hash table size to $2^{21}$ and the feature dimension to $4$. To improve the smoothness of the predicted normal vectors, we integrate standard frequency positional encoding as additional input for predicting surface normals:
\begin{equation}
    \gamma_\mathrm{f}(p)=(\sin{(2^k\pi p)},\cos{(2^k\pi p)})_{k=0}^{L-1}\,,
\end{equation}
where $L=2$ in our experiments. The material feature $\mathbf{f}_{\mathrm{mat}}$ is a $32$-dimensional vector.

The environment MLP $\mathcal{F}_{\mathrm{env}}$ is a $6$-layer MLP with hidden dimension $128$. It outputs a $32$-dimensional feature vector $\mathbf{f}_{\mathrm{env}}$. The diffuse MLP $\mathcal{F}_{\mathrm{d}}$ is a $2$-layer MLP with hidden dimension $32$, and the specular MLP $\mathcal{F}_{\mathrm{s}}$ is a $4$-layer MLP with hidden dimension $128$. The configuration of each MLP is determined experimentally to achieve a balance between training time and rendering quality.

\section{Training Details}
\label{sup:opti}
We train our model for $50$k iterations, with a batch size of $2^{19}$ sample points. Like Instant-NGP \cite{instant-ngp}, we use the Adam optimizer \cite{adam} with $\beta_1=0.9$, $\beta_2=0.99$, $\epsilon=10^{-15}$ for optimization. However, we employ distinct learning rate schedules for the hash grid and MLPs. Specifically, the learning rate for the hash grid logarithmically decays from $10^{-2}$ to $10^{-4}$, and the learning rate for MLPs logarithmically decays from $5\times 10^{-3}$ to $10^{-4}$ after a $5$k cosine warmup. The weight of the predicted normal loss $\mathcal{L}_\mathrm{n}$ logarithmically decays from $6\times 10^{-2}$ to $3\times 10^{-3}$ over first $20$k iterations. Furthermore, we employ a normalized weight decay of $10^{-2}$ for the hash grid, as introduced in Zip-NeRF \cite{zipnerf}. All experiments are conducted using Pytorch version 2.1.2 with CUDA 11.8, on a system equipped with a NVIDIA RTX 4090 GPU, running Ubuntu 22.04.4 as the operating system.

\section{Additional Ablation Studies}

\begin{table}[!t]
    \centering
    \begin{tabular}{l|ccc}
        \hline
        $\lambda_\mathrm{n}$ Settings & PSNR$\uparrow$    & SSIM$\uparrow$    & LPIPS$\downarrow$ \\
        \hline
        Full Model                    & \textbf{33.24}    & \textbf{0.971}    & \textbf{0.043}    \\
        No Reflect                    & 25.41             & 0.900             & 0.126             \\
        No Predicted Normal           & 25.90             & 0.906             & 0.126             \\
        No Frequency                  & \underline{32.58} & \underline{0.966} & \underline{0.050} \\
        No Hashgrid                   & $\mathit{30.41}$  & 0.949             & 0.073             \\
        No Grid Decay                 & 30.26             & $\mathit{0.956}$  & $\mathit{0.067}$  \\
        \hline
    \end{tabular}
    \caption{Ablation on architecture and regularization.}
    \label{tab:ablation_arch}
\end{table}

We conduct ablation studies on network architecture and regularization. Evaluation metrics on Glossy Synthetic \cite{nero} are presented in Table \ref{tab:ablation_arch}. We test the following settings: \textbf{No Reflect}, where the viewing direction is directly input into the environmental MLP $\mathcal{F}_\mathrm{env}$, without using its reflected direction; \textbf{No Predicted Normal}, where we compute reflection directions using the transmittance gradients directly, instead of relying on the predicted normals; \textbf{No Frequency}, where the frequency positional encoding $\gamma_\mathrm{f}$ is omitted from the input when predicting normal vectors; \textbf{No Hashgrid}, where we only use the frequency positional encoding to predict normal vectors; \textbf{No Grid Decay}, where the normalized weight decay is not applied to the hash grid.

\section{Performance Robustness}
Table \ref{tab:mean_std} presents the standard deviation of the our method across five independent runs, each optimized using a different random seed. These results substantiate the robustness of our model.

\section{Additional Results}
\begin{figure*}[htbp]
    \centering
    \includegraphics[width=\textwidth]{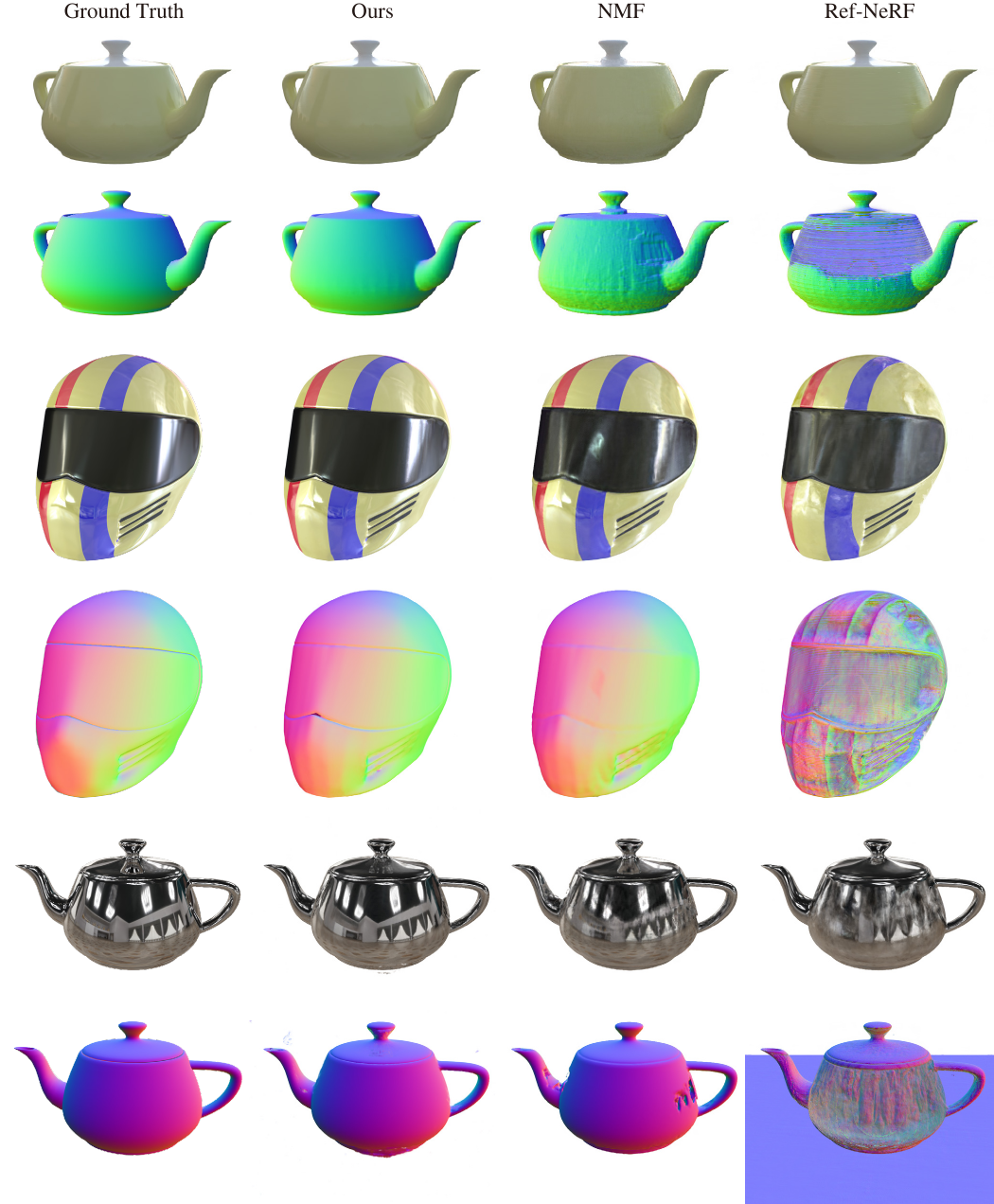}
    \caption{Visual comparisons with NeRF-based baselines, including NMF \cite{nmf} and Ref-NeRF \cite{ref-nerf}.}
    \label{fig:suppl_nerf}
\end{figure*}

\begin{figure*}[!t]
    \centering
    \includegraphics[width=\textwidth]{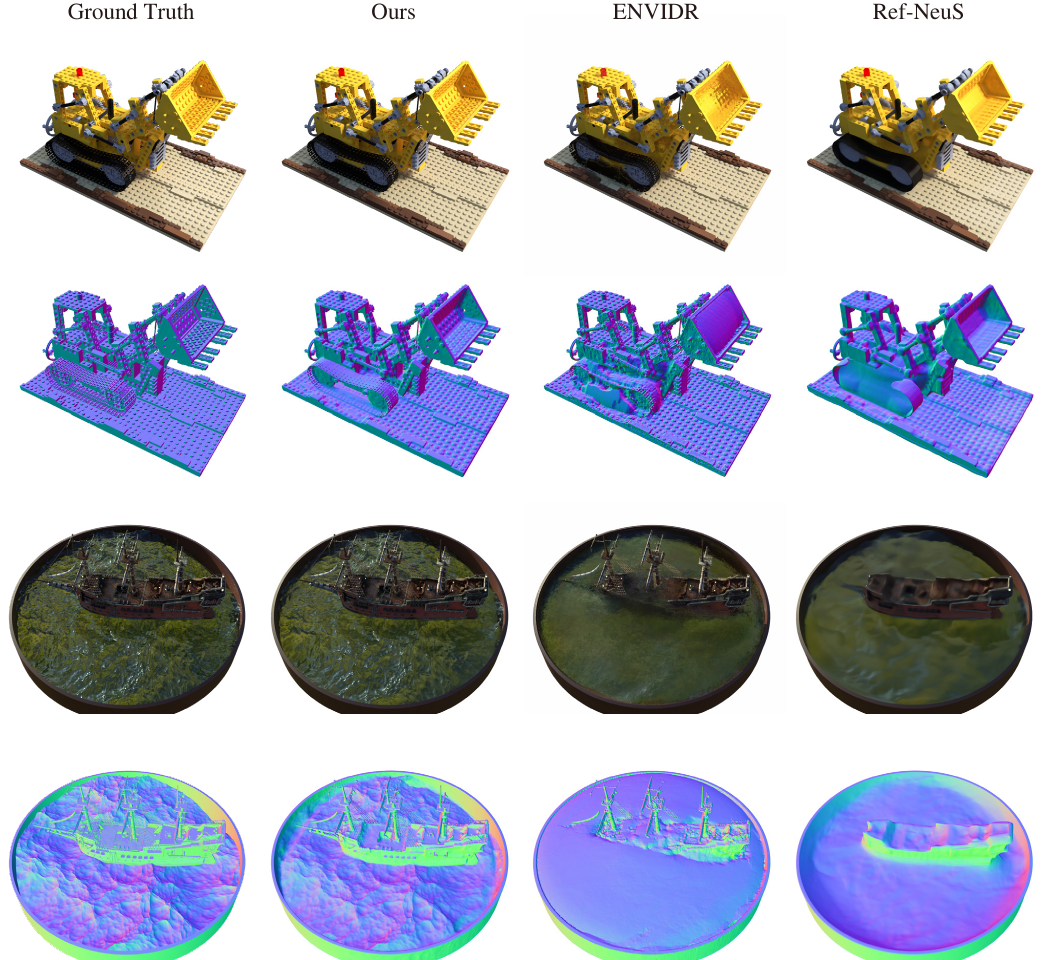}
    \caption{Visual comparisons with SDF-based baselines, including ENVIDR \cite{envidr} and Ref-NeuS \cite{ref-neus}.}
    \label{fig:suppl_sdf}
\end{figure*}

Figure \ref{fig:suppl_nerf} visualizes additional comparisons with NeRF-based baselines, and Figure \ref{fig:suppl_sdf} shows visual comparisons with SDF-based baselines. \Cref{tab:sota_blender,tab:sota_shiny,tab:sota_glossy,tab:sota_real} present per-scene evaluation metrics on NeRF Synthetic \cite{nerf}, Shiny Blender \cite{ref-nerf}, Glossy Synthetic \cite{nero} and real captured scenes from Ref-NeRF \cite{ref-nerf}. The results of Ref-NeRF \cite{ref-nerf} and NMF \cite{nmf} on NeRF Synthetic and Shiny Blender are extracted from their respective papers. We re-evaluate ENVIDR using the rendering images they released, employing the same code we used for computing metrics, to ensure the fairness of comparisons. Results on other datasets and results of other baselines are obtained by rerunning the official code released by these studies.

\begin{table*}[!t]
    \centering
    \begin{tabular}{l|cccccc|c}
        \hline
              & teapot      & toaster     & car         & ball        & coffee      & helmet      & Average     \\
        \hline
        PSNR  & 50.31±0.06  & 30.31±0.14  & 32.84±0.06  & 46.20±0.04  & 35.19±0.06  & 40.57±0.05  & 39.24±0.07  \\
        SSIM  & 0.999±0.000 & 0.961±0.001 & 0.971±0.000 & 0.994±0.000 & 0.976±0.000 & 0.993±0.000 & 0.982±0.000 \\
        LPIPS & 0.002±0.000 & 0.066±0.001 & 0.034±0.000 & 0.050±0.000 & 0.074±0.000 & 0.014±0.000 & 0.040±0.000 \\
        MAE   & 2.776±0.042 & 8.329±0.094 & 6.622±0.185 & 0.529±0.003 & 5.470±0.117 & 1.720±0.015 & 4.241±0.076 \\
        \hline
    \end{tabular}
    \caption{Mean and standard deviation of the performance of our method across five runs on Shiny Blender.}
    \label{tab:mean_std}
\end{table*}

\begin{table*}[t]
    \centering
    \begin{tabular}{lccccccccc}
        \hline
        \multicolumn{10}{c}{NeRF Synthetic \cite{nerf}}                                                                                                                                                      \\
                 & chair              & lego               & materials         & mic                & hotdog             & ficus              & drums              & ship               & Average            \\
        \hline
        \multicolumn{10}{c}{PSNR$\uparrow$}                                                                                                                                                                  \\
        \hline
        Zip-NeRF & $\mathit{35.77}$   & \underline{35.83}  & 31.02             & $\mathit{35.91}$   & \underline{38.01}  & \textbf{34.71}     & \underline{25.91}  & \textbf{32.35}     & $\mathit{33.69}$   \\
        Ref-NeRF & \underline{35.83}  & \textbf{36.25}     & \textbf{35.41}    & \underline{36.76}  & $\mathit{37.72}$   & $\mathit{33.91}$   & $\mathit{25.79}$   & $\mathit{30.28}$   & \underline{33.99}  \\
        NMF      & 32.27              & 32.98              & $\mathit{31.19}$  & 32.41              & 35.23              & 29.24              & 24.96              & 27.37              & 30.71              \\
        Ref-NeuS & 29.73              & 26.98              & 30.51             & 24.41              & 35.58              & 21.03              & 16.48              & 21.90              & 25.83              \\
        ENVIDR   & 33.45              & 29.21              & 29.43             & 32.39              & 30.14              & 30.38              & 24.71              & 24.96              & 29.33              \\
        Ours     & \textbf{36.59}     & $\mathit{35.80}$   & \underline{33.53} & \textbf{38.68}     & \textbf{38.57}     & \underline{34.03}  & \textbf{26.71}     & \underline{32.31}  & \textbf{34.53}     \\
        \hline\hline
        \multicolumn{10}{c}{SSIM$\uparrow$}                                                                                                                                                                  \\
        \hline
        Zip-NeRF & \underline{0.987}  & \textbf{0.983}     & $\mathit{0.968}$  & \underline{0.992}  & \textbf{0.987}     & \textbf{0.987}     & \textbf{0.950}     & \textbf{0.938}     & \textbf{0.974}     \\
        Ref-NeRF & $\mathit{0.984}$   & \underline{0.981}  & \textbf{0.983}    & \underline{0.992}  & $\mathit{0.984}$   & $\mathit{0.983}$   & \underline{0.937}  & $\mathit{0.880}$   & $\mathit{0.965}$   \\
        NMF      & 0.956              & $\mathit{0.963}$   & 0.959             & 0.977              & 0.964              & 0.952              & 0.917              & 0.828              & 0.940              \\
        Ref-NeuS & 0.942              & 0.923              & 0.967             & 0.960              & 0.973              & 0.892              & 0.861              & 0.813              & 0.916              \\
        ENVIDR   & 0.980              & 0.946              & 0.957             & $\mathit{0.983}$   & 0.950              & 0.973              & $\mathit{0.930}$   & 0.818              & 0.942              \\
        Ours     & \textbf{0.988}     & \underline{0.981}  & \underline{0.976} & \textbf{0.994}     & \underline{0.986}  & \underline{0.984}  & \textbf{0.950}     & \underline{0.921}  & \underline{0.973}  \\
        \hline\hline
        \multicolumn{10}{c}{LPIPS$\downarrow$}                                                                                                                                                               \\
        \hline
        Zip-NeRF & \underline{0.013}  & \textbf{0.015}     & 0.031             & \underline{0.006}  & \textbf{0.017}     & \textbf{0.013}     & \textbf{0.045}     & \textbf{0.082}     & \textbf{0.028}     \\
        Ref-NeRF & $\mathit{0.017}$   & $\mathit{0.018}$   & \textbf{0.022}    & $\mathit{0.007}$   & \underline{0.022}  & \underline{0.019}  & $\mathit{0.059}$   & $\mathit{0.139}$   & $\mathit{0.038}$   \\
        NMF      & 0.044              & 0.024              & \underline{0.026} & 0.022              & 0.046              & 0.044              & 0.068              & 0.149              & 0.053              \\
        Ref-NeuS & 0.065              & 0.092              & 0.038             & 0.044              & $\mathit{0.042}$   & 0.106              & 0.169              & 0.238              & 0.099              \\
        ENVIDR   & 0.020              & 0.053              & 0.041             & 0.018              & 0.072              & $\mathit{0.033}$   & 0.071              & 0.200              & 0.064              \\
        Ours     & \textbf{0.012}     & \underline{0.017}  & $\mathit{0.028}$  & \textbf{0.005}     & \textbf{0.017}     & \underline{0.019}  & \underline{0.053}  & \underline{0.092}  & \underline{0.030}  \\
        \hline\hline
        \multicolumn{10}{c}{MAE$\downarrow$}                                                                                                                                                                 \\
        \hline
        Zip-NeRF & -                  & -                  & -                 & -                  & -                  & -                  & -                  & -                  & -                  \\
        Ref-NeRF & 19.852             & \underline{24.469} & 9.531             & 24.938             & 13.211             & 41.052             & 27.853             & 31.707             & 24.077             \\
        NMF      & 14.330             & $\mathit{25.918}$  & $\mathit{8.101}$  & 20.144             & \textbf{10.043}    & $\mathit{37.405}$  & \underline{21.524} & 30.152             & $\mathit{20.952}$  \\
        Ref-NeuS & \underline{11.056} & 28.302             & \underline{7.968} & $\mathit{19.047}$  & $\mathit{12.215}$  & 44.354             & 36.630             & \underline{24.214} & 22.973             \\
        ENVIDR   & $\mathit{11.327}$  & 28.290             & 8.374             & \underline{17.260} & 23.831             & \textbf{24.551}    & $\mathit{22.198}$  & $\mathit{28.560}$  & \underline{20.549} \\
        Ours     & \textbf{9.348}     & \textbf{20.261}    & \textbf{7.555}    & \textbf{14.124}    & \underline{11.487} & \underline{26.935} & \textbf{15.180}    & \textbf{10.998}    & \textbf{14.486}    \\
        \hline
    \end{tabular}
    \caption{Per-scene quantitative results on NeRF Synthetic \cite{nerf} dataset. The best results are \textbf{bold}, the second best results are \underline{underlined}, and the third best results are \textit{italics}.}
    \label{tab:sota_blender}
\end{table*}

\begin{table*}[t]
    \centering
    \begin{tabular}{lccccccc}
        \hline
        \multicolumn{8}{c}{Shiny Blender \cite{ref-nerf}}                                                                                                    \\
                 & teapot            & toaster           & car               & ball              & coffee            & helmet            & Average           \\
        \hline
        \multicolumn{8}{c}{PSNR$\uparrow$}                                                                                                                   \\
        \hline
        Zip-NeRF & $\mathit{46.10}$  & 24.27             & 27.32             & 25.37             & 30.61             & 27.03             & 30.12             \\
        Ref-NeRF & \underline{47.90} & 25.70             & \underline{30.82} & \textbf{47.46}    & \underline{34.21} & 29.68             & \underline{35.96} \\
        NMF      & 45.29             & $\mathit{27.52}$  & $\mathit{30.28}$  & 38.41             & 31.47             & 34.38             & 34.56             \\
        Ref-NeuS & 37.14             & \underline{28.50} & 28.14             & 37.42             & 32.99             & $\mathit{35.34}$  & 33.26             \\
        ENVIDR   & 43.90             & 26.41             & 29.50             & $\mathit{41.13}$  & $\mathit{33.63}$  & \underline{35.56} & $\mathit{35.02}$  \\
        Ours     & \textbf{50.31}    & \textbf{30.31}    & \textbf{32.84}    & \underline{46.20} & \textbf{35.19}    & \textbf{40.57}    & \textbf{39.24}    \\
        \hline\hline
        \multicolumn{8}{c}{SSIM$\uparrow$}                                                                                                                   \\
        \hline
        Zip-NeRF & $\mathit{0.997}$  & 0.918             & 0.933             & 0.924             & 0.966             & 0.948             & 0.948             \\
        Ref-NeRF & \underline{0.998} & 0.922             & $\mathit{0.955}$  & \textbf{0.995}    & \underline{0.974} & 0.958             & 0.967             \\
        NMF      & 0.996             & 0.917             & 0.951             & 0.983             & 0.960             & 0.969             & 0.963             \\
        Ref-NeuS & 0.996             & \underline{0.939} & 0.945             & $\mathit{0.992}$  & 0.969             & \underline{0.986} & $\mathit{0.971}$  \\
        ENVIDR   & \underline{0.998} & $\mathit{0.930}$  & \underline{0.957} & $\mathit{0.992}$  & $\mathit{0.971}$  & $\mathit{0.985}$  & \underline{0.972} \\
        Ours     & \textbf{0.999}    & \textbf{0.961}    & \textbf{0.971}    & \underline{0.994} & \textbf{0.976}    & \textbf{0.993}    & \textbf{0.982}    \\
        \hline\hline
        \multicolumn{8}{c}{LPIPS$\downarrow$}                                                                                                                \\
        \hline
        Zip-NeRF & 0.007             & 0.097             & 0.055             & 0.201             & 0.089             & 0.087             & 0.089             \\
        Ref-NeRF & \underline{0.004} & $\mathit{0.095}$  & \underline{0.041} & 0.059             & $\mathit{0.078}$  & 0.075             & 0.059             \\
        NMF      & 0.010             & 0.104             & \textbf{0.034}    & \underline{0.046} & \textbf{0.069}    & 0.055             & \underline{0.053} \\
        Ref-NeuS & 0.009             & \underline{0.076} & 0.060             & 0.058             & 0.093             & \underline{0.023} & $\mathit{0.053}$  \\
        ENVIDR   & $\mathit{0.006}$  & 0.113             & $\mathit{0.049}$  & \textbf{0.037}    & 0.080             & $\mathit{0.034}$  & 0.053             \\
        Ours     & \textbf{0.002}    & \textbf{0.066}    & \textbf{0.034}    & $\mathit{0.050}$  & \underline{0.074} & \textbf{0.014}    & \textbf{0.040}    \\
        \hline\hline
        \multicolumn{8}{c}{MAE$\downarrow$}                                                                                                                  \\
        \hline
        Zip-NeRF & -                 & -                 & -                 & -                 & -                 & -                 & -                 \\
        Ref-NeRF & 9.234             & 42.870            & 14.927            & 1.548             & 12.240            & 29.484            & 18.384            \\
        NMF      & 5.672             & $\mathit{6.660}$  & $\mathit{7.742}$  & 0.723             & 13.173            & 2.395             & 6.061             \\
        Ref-NeuS & $\mathit{3.510}$  & \textbf{5.560}    & 7.867             & \underline{0.475} & $\mathit{10.018}$ & \textbf{1.683}    & $\mathit{4.852}$  \\
        ENVIDR   & \textbf{2.634}    & \underline{6.496} & \underline{7.193} & \textbf{0.245}    & \underline{9.300} & $\mathit{1.744}$  & \underline{4.602} \\
        Ours     & \underline{2.776} & 8.329             & \textbf{6.622}    & $\mathit{0.529}$  & \textbf{5.470}    & \underline{1.720} & \textbf{4.241}    \\
        \hline
    \end{tabular}
    \caption{Per-scene quantitative results on Shiny Blender \cite{ref-nerf} dataset. The best results are \textbf{bold}, the second best results are \underline{underlined}, and the third best results are \textit{italics}.}
    \label{tab:sota_shiny}
\end{table*}

\begin{table*}[t]
    \centering
    \begin{tabular}{lccccccccc}
        \hline
        \multicolumn{10}{c}{Glossy Synthetic \cite{nero}}                                                                                                                                            \\
                 & angel             & bell              & cat               & horse             & luyu              & potion            & tbell             & utah\_teapot      & Average           \\
        \hline
        \multicolumn{10}{c}{PSNR$\uparrow$}                                                                                                                                                          \\
        \hline
        Zip-NeRF & \textbf{30.28}    & 28.69             & 35.38             & \underline{30.27} & \underline{29.91} & 33.41             & 27.48             & 23.64             & 29.88             \\
        Ref-NeRF & 23.36             & 30.24             & \underline{37.31} & $\mathit{29.49}$  & $\mathit{29.70}$  & $\mathit{34.63}$  & \underline{30.43} & \textbf{27.94}    & $\mathit{30.39}$  \\
        NMF      & $\mathit{28.94}$  & $\mathit{30.67}$  & 32.65             & 25.93             & 28.68             & 29.46             & 26.97             & 26.77             & 28.76             \\
        Ref-NeuS & \underline{29.53} & \underline{35.35} & $\mathit{36.04}$  & 25.89             & 28.32             & \underline{35.24} & 26.06             & \underline{27.46} & \underline{30.49} \\
        ENVIDR   & 27.88             & 29.82             & 30.12             & 28.62             & 26.33             & 29.91             & $\mathit{28.85}$  & 26.24             & 28.47             \\
        Ours     & 28.36             & \textbf{37.09}    & \textbf{38.37}    & \textbf{31.55}    & \textbf{31.70}    & \textbf{37.40}    & \textbf{34.13}    & $\mathit{27.30}$  & \textbf{33.24}    \\
        \hline\hline
        \multicolumn{10}{c}{SSIM$\uparrow$}                                                                                                                                                          \\
        \hline
        Zip-NeRF & \underline{0.945} & 0.929             & $\mathit{0.978}$  & \underline{0.974} & \underline{0.944} & $\mathit{0.959}$  & 0.930             & 0.896             & 0.944             \\
        Ref-NeRF & 0.875             & 0.938             & \underline{0.982} & 0.964             & $\mathit{0.929}$  & 0.957             & 0.947             & 0.934             & 0.941             \\
        NMF      & 0.925             & 0.944             & 0.959             & 0.950             & 0.924             & 0.912             & 0.911             & $\mathit{0.935}$  & 0.933             \\
        Ref-NeuS & $\mathit{0.933}$  & \underline{0.975} & \textbf{0.987}    & 0.943             & 0.923             & \underline{0.964} & \underline{0.965} & \textbf{0.957}    & \underline{0.956} \\
        ENVIDR   & 0.929             & $\mathit{0.964}$  & 0.960             & $\mathit{0.967}$  & 0.915             & 0.933             & $\mathit{0.957}$  & \textbf{0.957}    & $\mathit{0.948}$  \\
        Ours     & \textbf{0.952}    & \textbf{0.982}    & \textbf{0.987}    & \textbf{0.980}    & \textbf{0.958}    & \textbf{0.979}    & \textbf{0.976}    & \underline{0.954} & \textbf{0.971}    \\
        \hline\hline
        \multicolumn{10}{c}{LPIPS$\downarrow$}                                                                                                                                                       \\
        \hline
        Zip-NeRF & \textbf{0.055}    & 0.092             & $\mathit{0.036}$  & \textbf{0.021}    & \underline{0.053} & \underline{0.068} & 0.095             & 0.101             & $\mathit{0.065}$  \\
        Ref-NeRF & 0.123             & 0.083             & \underline{0.034} & 0.034             & $\mathit{0.068}$  & $\mathit{0.081}$  & 0.091             & 0.078             & 0.074             \\
        NMF      & 0.076             & 0.083             & 0.069             & 0.043             & $\mathit{0.068}$  & 0.124             & 0.143             & 0.079             & 0.086             \\
        Ref-NeuS & 0.074             & \underline{0.050} & \textbf{0.024}    & 0.056             & 0.086             & 0.082             & \underline{0.067} & $\mathit{0.071}$  & \underline{0.064} \\
        ENVIDR   & $\mathit{0.070}$  & $\mathit{0.061}$  & 0.057             & $\mathit{0.029}$  & 0.070             & 0.096             & $\mathit{0.081}$  & \textbf{0.060}    & 0.066             \\
        Ours     & \underline{0.058} & \textbf{0.034}    & \textbf{0.024}    & \underline{0.024} & \textbf{0.044}    & \textbf{0.040}    & \textbf{0.051}    & \underline{0.067} & \textbf{0.043}    \\
        \hline
    \end{tabular}
    \caption{Per-scene quantitative results on Glossy Synthetic \cite{nero} dataset. The best results are \textbf{bold}, the second best results are \underline{underlined}, and the third best results are \textit{italics}.}
    \label{tab:sota_glossy}
\end{table*}

\begin{table*}[t]
    \centering
    \begin{tabular}{lcccc}
        \hline
        \multicolumn{5}{c}{Real Captured Scenes \cite{ref-nerf}}                                 \\
                 & sedan             & toycar            & gardenspheres     & Average           \\
        \hline
        \multicolumn{5}{c}{PSNR$\uparrow$}                                                       \\
        \hline
        Zip-NeRF & \textbf{25.85}    & $\mathit{23.29}$  & $\mathit{21.17}$  & $\mathit{23.44}$  \\
        Ref-NeRF & $\mathit{25.20}$  & \textbf{24.40}    & \underline{22.57} & \underline{24.06} \\
        Ours     & \underline{25.47} & \underline{24.15} & \textbf{22.66}    & \textbf{24.09}    \\
        \hline\hline
        \multicolumn{5}{c}{SSIM$\uparrow$}                                                       \\
        \hline
        Zip-NeRF & \textbf{0.732}    & $\mathit{0.621}$  & \underline{0.524} & \underline{0.626} \\
        Ref-NeRF & $\mathit{0.639}$  & \underline{0.627} & $\mathit{0.502}$  & $\mathit{0.589}$  \\
        Ours     & \underline{0.682} & \textbf{0.647}    & \textbf{0.562}    & \textbf{0.630}    \\
        \hline\hline
        \multicolumn{5}{c}{LPIPS$\downarrow$}                                                    \\
        \hline
        Zip-NeRF & \textbf{0.258}    & \textbf{0.247}    & \textbf{0.265}    & \textbf{0.257}    \\
        Ref-NeRF & $\mathit{0.406}$  & $\mathit{0.292}$  & $\mathit{0.366}$  & $\mathit{0.355}$  \\
        Ours     & \underline{0.328} & \underline{0.268} & \underline{0.283} & \underline{0.293} \\
        \hline
    \end{tabular}
    \caption{Per-scene quantitative results on real captured scenes from Ref-NeRF \cite{ref-nerf} dataset. The best results are \textbf{bold}, the second best results are \underline{underlined}, and the third best results are \textit{italics}.}
    \label{tab:sota_real}
\end{table*}

\end{document}